\documentclass[letterpaper, 10 pt, conference]{ieeeconf}

\UseRawInputEncoding
\IEEEoverridecommandlockouts
\usepackage[letterpaper,top=54pt,bottom=54pt,left=54pt,right=54pt]{geometry}

\usepackage{times}
\usepackage{amsmath,amssymb,amsopn,amstext,amsfonts}
\usepackage{cancel}
\usepackage[space]{cite}
\usepackage{balance}
\usepackage{mathtools}
\usepackage[ruled,vlined,linesnumbered]{algorithm2e}
\SetAlgoNlRelativeSize{-1}
\SetAlCapNameFnt{\small}
\SetAlCapFnt{\small}
\SetKwFor{ParallelForAll}{forall}{do in parallel}{end for}

\usepackage{bm}

\usepackage{diagbox}
\usepackage{float}
\usepackage{pifont}
\usepackage{amsmath}
\usepackage{multirow}
\usepackage{booktabs}
\usepackage{url}
\usepackage{svg}
\usepackage{threeparttable}
\usepackage[hidelinks]{hyperref}
\usepackage{makecell}

\usepackage{array}
\usepackage{marvosym}

\DeclareGraphicsExtensions{.png,.jpg,.eps,.pdf}
\DeclareGraphicsExtensions{.pdf}

\title{\LARGE \bf 
Neural-Primitive: An Efficient End-to-end Local Planner with Primitive-based Imitation Learning for Autonomous Flight}

\author{Zhitao Liu\textsuperscript{1,2,3,*},
Guangtong Xu\textsuperscript{4,*},
Zihan Wang\textsuperscript{5},
Jialiang Hou\textsuperscript{1,6,\textdagger},
Chao Xu\textsuperscript{1,2,\textdagger},
and Fei Gao\textsuperscript{1,6,\textdagger}%
\thanks{\textsuperscript{*}Indicates equal contribution.}%
\thanks{\textsuperscript{\textdagger}Corresponding Authors: Jialiang Hou; Fei Gao; Chao Xu.}%
\thanks{This work was supported by the National Key R\&D Program of China under Grant No. 2023YFB4706600, the Zhejiang Provincial Science and Technology Plan Project under Grant No. 2024C01170, and the National Natural Science Foundation of China under Grant No. 62322314 and No. 62203256.}%
\thanks{\textsuperscript{1}Institute of Cyber-Systems and Control, College of Control Science and Engineering, Zhejiang University, Hangzhou 310027, China.}%
\thanks{\textsuperscript{2}Huzhou Institute, Zhejiang University, Huzhou 313000, China.}%
\thanks{\textsuperscript{3}Institute of Systems Engineering, China Academy of Engineering Physics, Mianyang 621999, China.}%
\thanks{\textsuperscript{4}School of Automation, Hangzhou Dianzi University, Hangzhou 310018, China.}%
\thanks{\textsuperscript{5}Department of Automation, North China Electric Power University (Baoding), Baoding 071003, China.}%
\thanks{\textsuperscript{6}Differential Robotics Technology Company, Hangzhou 311121, China.}%
\thanks{\raggedright E-mail: zhitaoliu@zju.edu.cn; jlhou25@zju.edu.cn; fgaoaa@zju.edu.cn; cxu@zju.edu.cn.}%
\thanks{Our project page with videos is at \url{https://ZhitaoLiu.github.io/neural-primitive/}.}}

\let\oldtwocolumn\twocolumn
\renewcommand\twocolumn[1][]{
	\oldtwocolumn[{#1}{
        \vspace{-0.3cm}
		\begin{center}        
            \includegraphics[width=\textwidth]{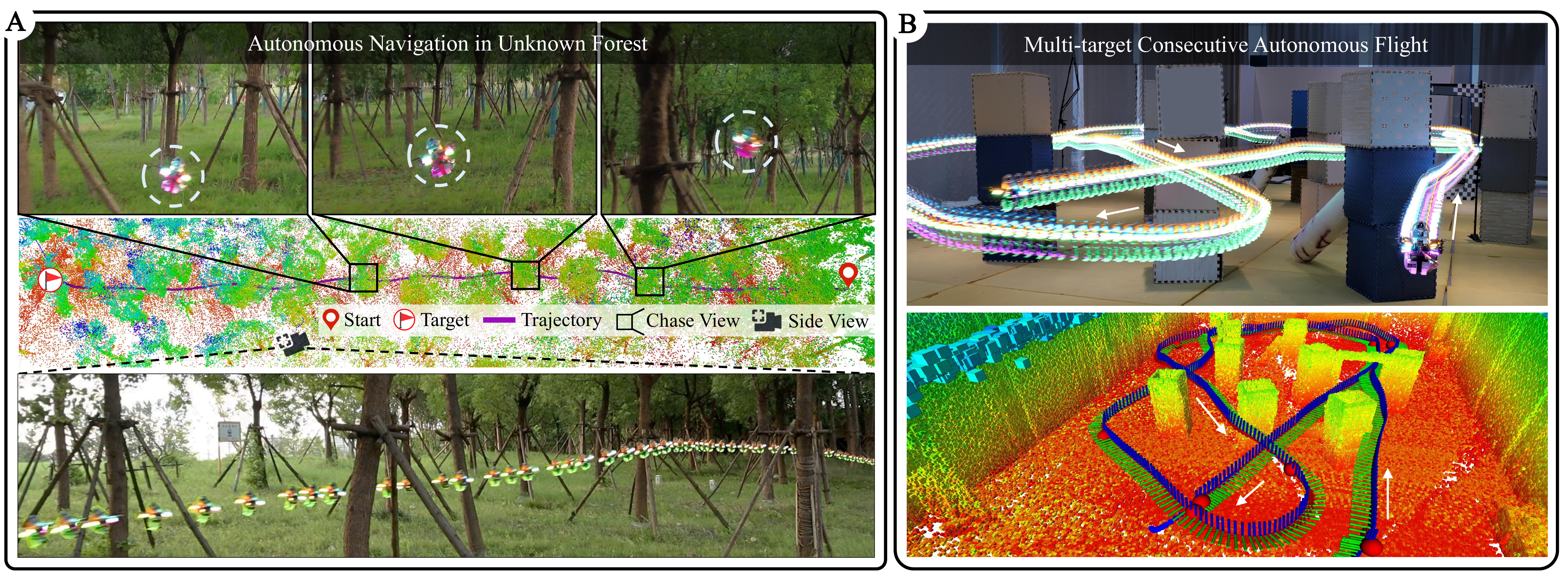}
            \vskip 4pt
            \begin{minipage}{1.0\textwidth} 
                \footnotesize
                \refstepcounter{figure} 
                \label{fig:Real_world_exp}
                Fig.~\thefigure. \textbf{Real-world experiments with zero-shot network deployment.}
                \textbf{(A)} The quadrotor autonomously navigates through a dense cluttered forest and reaches the target 60m ahead with top-level performance. 
                \textbf{(B)} Trajectory of multi-target consecutive autonomous flight in confined indoor space with random obstacles.
            \end{minipage}
		\end{center}
	}]
}

\begin{document}

\maketitle

\begin{abstract}

Autonomous flight in unknown cluttered environments is hindered by the \textit{computation-quality-memory} trilemma of onboard trajectory generation. In this paper, we propose an efficient end-to-end local planner via imitation learning. A lightweight offline-primitive-based dataset collection framework is designed to produce safe and high-quality trajectory primitives in non-convex environments. A compact neural network directly maps sensory inputs to polynomial coefficients that inherently encode higher-order dynamical information. The learned policy generates smooth, {empirically collision-free and dynamically feasible} trajectories in real time without back-end solving. It achieves ultra-fast computation (below 1ms on a standard desktop and average 3.68ms during onboard flight), while maintaining low onboard memory requirements (less than 1.5MiB). Extensive simulation benchmarks demonstrate superiority in both planning latency and target-reaching progress quality. Zero-shot deployment in real-world experiments further validates the robust sim-to-real transfer capability of the proposed method.

\end{abstract}
\section{Introduction}
\label{sec:introduction}

Autonomous flight in unknown cluttered environments using only onboard noisy sensing and resource-constrained computation, epitomizes one of the grand challenges  for unmanned aerial vehicles (UAVs) \cite{yang2018grand}. Despite remarkable progress \cite{loquercio2021learning,ren2025safety,zhou2022swarm}, fast generation of dynamically feasible, collision-free, and efficiently target-reaching trajectories still remains open.

Numerous trajectory generation approaches \cite{zhou2020ego,Lu2025FAPP,wang2025unlocking,zhou2019robust} couple obstacle avoidance into joint optimization through strict constraints or weighted penalties. However, simultaneously considering excessive factors forces trade-offs, making solutions prone to local minima or even intractable, degrading overall trajectory quality. In contrast, motion primitive-based methods \cite{hou2025primitive,ryll2019efficient,Dharmadhikari2020Motion,zhang2020falco} decouple obstacle avoidance out through a stepwise pipeline of free-space sampling, collision removal, and optimal selection. This facilitates near-optimal solutions but renders performance hinging on the primitive library: online-generated methods struggle to produce both dynamically feasible and diverse primitives in a short time; whereas offline-generated methods can pre-compute high-quality ones without latency concerns, yet suffer from higher-order state discontinuities during online concatenation, as the fixed library cannot cover all required initial states with limited onboard memory. Furthermore, these approaches remain subject to processing latency and compounding errors due to hierarchical planning frameworks. Recently, learning-based planning has gained significant attention \cite{han2025hierarchically,lu2024you,wu2024deep}. Although claiming end-to-end planning capability, most still require online back-end optimization \cite{han2025dyna}, closed-form solving \cite{lu2024you}, or projection \cite{loquercio2021learning} after network inference to yield controller-executable trajectories, ultimately restricting achievable planning efficiency. These methods invariably face a \textit{computation-quality-memory} trilemma, namely the inability to simultaneously satisfy fast online computation, near-optimal target-reaching trajectories, and low onboard memory consumption. 

To address the aforementioned challenges, we propose \textit{Neural-Primitive}, an efficient end-to-end local planner realized by imitation learning of a customized offline primitive strategy. Unlike naive replication of existing planners, which is limited by expert performance, we inherit the core idea from offline primitive library-based methods, namely generating high-quality primitives through sampling-exploration and obstacle-avoidance-decoupling. At the same time, we resolve their major drawbacks (high-order discontinuities and storage burden of large libraries) by employing online inference with a compact neural network (memory size less than 1.5MiB). We design a primitive-strategy-embedded dataset collection procedure that mitigates covariate shift \cite{damanik2024lics} without resorting to resource-intensive approaches such as DAgger \cite{tejaswi2022constrained}. This enables subsequent training to capture precise target-reaching behaviors in cluttered environments. Compared with existing learning-based planners, our policy directly predicts polynomial coefficients from sensory inputs, intrinsically encoding high-order dynamical information and producing {practically} controller-executable trajectories without back-end solving\footnote{{``Without back-end solving'' means that no additional optimization, quadratic programming, or boundary-value problem solving is needed after network inference.}}. This design results in a highly integrated end-to-end planning scheme. Together with a streamlined network architecture that avoids convolutions and other costly operations, the proposed method achieves ultra-fast computation ($<$ 1ms on standard desktop and mean 3.68ms onboard) while maintaining consistency across scenarios. {To bridge the sim-to-real gap from sensory mismatch, we design a point cloud preprocessing technique that adapts to obstacle density variations. This boosts robustness and generalization atop domain randomization.} Consequently, our method achieves fast online generation of near-straight trajectories toward the target across unknown cluttered environments. 

We benchmark with representative state-of-the-art local planners \cite{zhou2020ego,zhou2019robust,ren2025safety,lu2024you} through extensive simulations. Statistical results demonstrate superior performance in computation time, success rate, flight length, and flight time. The generalization and sim-to-real capabilities of the proposed method are further validated in complex previously unseen maps and real-world experiments. The contributions of this paper are summarized as follows:
\begin{enumerate}
    \item A lightweight offline-primitive-based dataset collection framework, which efficiently produces safe and high-quality trajectory primitives in non-convex environments.
    \item A compact imitation learning neural network, which can output polynomial coefficients {empirically satisfying safety, dynamical feasibility}, and target-reaching progress quality under the supervision of the expert dataset, with enhanced generalization capability enabled by the proposed point cloud pre-processing training technique.
    \item An ultra-fast and high-quality end-to-end local planner, which directly maps sensory inputs to {high-order continuous} polynomial trajectories without back-end solving.
    \item Comprehensive simulation benchmarks and zero-shot deployment in real-world experiments are conducted {to validate} the superiority of our method.
\end{enumerate}

\section{Related Work}
\label{sec:related_work}

\begin{figure*}[t] 
    \centering
    \includegraphics[width=\textwidth]{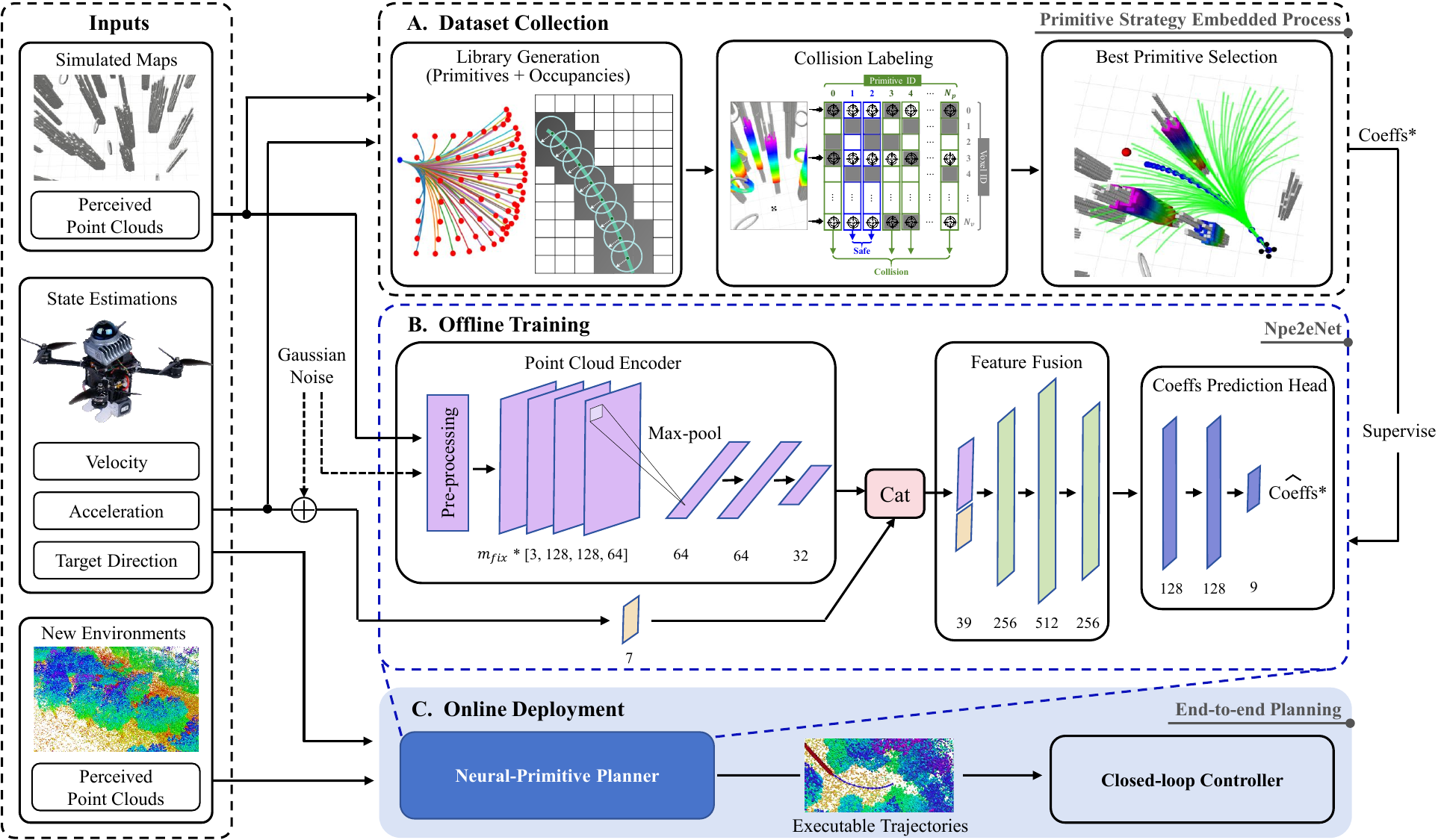}
    \vspace{-0.3cm}
    \caption{\textbf{System overview.} The  end-to-end planner generates smooth, {empirically collision-free and dynamically feasible} trajectories directly from onboard sensory inputs. \textbf{(A)} Training datasets are collected entirely in simulation using a customized primitive strategy. \textbf{(B)} A neural  network learns to predict polynomial coefficients that inherently encode high-order dynamical information for direct execution by the low-level controller. \textbf{(C)} The learned policy is zero-shot deployed for fast online inference in previously unseen environments without fine-tuning.}
    \label{fig:System overview}
    \vspace{-0.5cm}
\end{figure*}

\subsection{Primitive-based Motion Planning for UAV}

Primitive-based methods cast motion planning as primitive selection, enhancing solution space multimodality through free-space sampling and obstacle-avoidance decoupling. Based on primitive library generation, they can be categorized into online and offline approaches. In \cite{ryll2019efficient,yang2021intention,Lee2024BPMP}, polynomial trajectory candidates are obtained online by sampling terminal states and solving boundary value problems (BVPs) in closed form. However, dynamical feasibility is not considered during generation and must be verified individually through post-checking \cite{mueller2015computationally}. Some works \cite{yang2019online,collins2020efficient,Dharmadhikari2020Motion} sample control inputs and generate primitives online through forward calculation with simplified motion models, which can accommodate input limits but degrade trajectory quality due to model inaccuracy. Online methods can hardly generate trajectory primitives with sufficient diversity and quality within a short planning horizon. Conversely, studies \cite{zhang2018p,zhang2020falco} generate plenty of path primitives offline for online selection, but geometric paths neglect dynamical properties (velocity, acceleration, {etc.}), thus fail to ensure higher-order continuity during online concatenation. Hou et al. \cite{hou2025primitive} further {generate} time-optimal trajectory primitives offline with dense initial velocity discretizations to promote velocity continuity, {but acceleration continuity still remains unresolved.} In practice, although offline methods offer broad primitive coverage regardless of latency concerns, {storing} all high-order initial states for online concatenation without exceeding onboard memory is nearly infeasible (a rough estimate based on \cite{hou2025primitive} indicates that covering initial states up to 3m/s and 3m/s$^2$ with 0.1 steps would require far more than 1000GB of memory), leaving discontinuity an inherent unresolved drawback.

\subsection{Learning-based Motion Planning for UAV}
Neural networks, with their strong capacity for environmental abstraction and rapid online inference, show promising potential in enhancing UAV motion planning efficiency. {Wu et al. \cite{wu2024deep} use a network to predict time allocations for piecewise trajectories, but still rely on the traditional hierarchical pipeline of mapping, path finding, and trajectory optimization, leading to high computational cost.} In \cite{han2025dyna,han2025hierarchically}, networks replace mapping and path finding, but online optimization is still required for generating executable trajectories. {The network in \cite{lu2024you} predicts end-state terms for pre-sampled primitives and solves one online via a closed-form BVP back-end. However, dynamical process constraints are not explicitly enforced, and its coupled training objective introduces inherent trade-offs, making solutions prone to local minima.} Loquercio et al. \cite{loquercio2021learning} utilize {a} network to output waypoints, but they contain only geometric information without dynamical properties, requiring real-time projection after network inference to obtain polynomial trajectories. Overall, existing learning-based planning methods have yet to fully demonstrate end-to-end planning without back-end trajectory solving, and direct generation of controller-executable trajectories with efficient target-reaching progress quality still remains improvable.

\section{System Overview}
\label{sec:system_overview}

Fig. \ref{fig:System overview} illustrates the three-stage pipeline of the proposed system.

Firstly, all training datasets are collected in simulation, {incorporating a customized primitive strategy to strengthen the dynamical feasibility, collision avoidance, and task-oriented trajectory quality of the expert primitives.} The resulting primitive from each successful replanning step is stored along with the perceived point clouds, drone states, and target direction. Instead of saving path points that only represent geometric positions \cite{loquercio2021learning}, the polynomial coefficients of the parameterized primitive are recorded, which inherently encode high-order dynamical information.

Subsequently, a policy network based on multilayer perceptrons (MLPs) is trained offline in a supervised manner utilizing the pre-collected datasets. It learns to generate continuous primitives from discrete samples, predicting polynomial coefficients from sensory inputs. This addresses the velocity or acceleration discontinuity issue inherent in traditional offline primitive library-based planning methods \cite{hou2025primitive,zhang2020falco,zhang2018p}, while also eliminating the need for back-end trajectory solving \cite{lu2023lpnet,lu2024you,han2025dyna,wu2024deep,han2025hierarchically} or projection \cite{loquercio2021learning} required in existing learning-based planning approaches, thereby enabling an end-to-end planning framework. To reduce the sensory gap for sim-to-real transfer, point clouds are preprocessed prior to formal training and augmented with Gaussian noise, which is also injected into other inputs to enhance generalization.

Finally, the learned policy is zero-shot deployed onboard for inference in previously unseen environments without any fine-tuning. It directly maps onboard sensory data to {practically} executable trajectories in a receding-horizon fashion, which are sent to the low-level controller for execution.
\section{Methodology}
\label{sec:methodology}

\subsection{Primitive Strategy Embedded Dataset Collection}
\subsubsection{Collection Procedure}

Dataset generation for imitation learning needs to account for the covariate shift \cite{damanik2024lics} arising from distribution mismatch between expert demonstrations and real-world deployment. Rather than relying on resource-intensive methods like DAgger \cite{tejaswi2022constrained}, we collect the full dataset prior to training. Specifically, data are obtained in a closed-loop manner, recording only those from the successful completion of randomly initialized navigation tasks. Evidently, replanning can be triggered at arbitrary initial states throughout the process. However, constructing a large offline library covering all the states for selection is infeasible due to overwhelming memory and computation burden. Therefore, we propose a dataset collection procedure embedded with a customized primitive strategy, as detailed in Algorithm \ref{alg:Dataset collection}.

\begin{algorithm}[t]
	\DontPrintSemicolon
	\caption{Dataset Collection with Primitive Strategy}
	\label{alg:Dataset collection}
	\KwIn{$n_{max}$, $n_{thrd}$, $\Theta_{map}$, $\Theta_{init}$, $\Theta_{lib}$}
	\KwOut{$\mathcal{D}$}
	
	$\mathcal{D} \gets \emptyset$, $n_{valid} \gets 0$\;
	\While{$n_{valid} < n_{max}$}{
		$\mathcal{M} \gets randomMapGen(\Theta_{map})$
		
		\ParallelForAll{$ thrd_i \in \{1, \dots, n_{thrd}\}$ }{
			$\bm{s}_{0}, \bm{p}_{target}, \bm{d}_{0} \gets randomInitGen(\Theta_{init})$
			
			$\mathcal{D}_{temp}.clear()$
			
			\While{$\mathrm{true}$}{
				$\mathcal{L}.clear()$
				
				\ForAll{$\bm{s}_{f} \in endStateSet(\Theta_{lib})$}{
					$\bm{\xi} \gets solveQP(\bm{s}_{0}, \bm{s}_{f}, \tau, \Theta_{lib})$
					
					$\mathcal{O} \gets sampleTraversalVoxels(\bm{\xi})$
					
					$\mathcal{L}.pushback(prim(\bm{\xi}, \mathcal{O}))$
				}
				
				$\mathcal{P} \gets sensePointCloud(\bm{s}_{0}, \mathcal{M})$
				
				$\mathcal{L}_{safe} \gets collisionLabeling(\mathcal{L}, \mathcal{P}, R^V_W)$
				
				$prim^* \gets selectBest(\mathcal{L}_{safe}, \bm{p}_{target}, R^V_W)$
				
				\If{$prim^* \ \mathrm{is \ empty}$}{
					\textbf{break}
				}
				
				$\bm{c}^* \gets prim^*.getCoeffs()$
				
				$\bm{v}_{0}, \bm{a}_{0} \gets \bm{s}_{0}.getVelAcc()$
				
				$\mathcal{D}_{temp}.pushback(data(\bm{v}_{0}, \bm{a}_{0}, \bm{d}_{0}, \mathcal{P}, \bm{c}^*))$
				
				$\bm{s}_{0}, \bm{d}_{0} \gets executeTraj(prim^*)$
				
				\If{$targetReach(\bm{s}_{0}, \bm{p}_{target})$}{
					$\mathcal{D} \gets \mathcal{D} \cup \mathcal{D}_{temp}$
					
					$n_{valid} \gets n_{valid} + \mathcal{D}_{temp}.size()$
					
					\textbf{break}
				}
			}
			
			\If{$n_{valid} \ge n_{max}$}{
				terminate all threads
			}
		}
	}
	\Return{$\mathcal{D}$}
\end{algorithm}

A randomized simulation map $\mathcal{M}$ is first generated from given parameters $\Theta_{map}$ (Line 3). Then up to {$n_{thrd}$} navigation tasks run in multithreaded parallel to accelerate the process. Each task randomly initializes the target position $\bm{p}_{target}$ and drone state $\bm{s}_{0}$ based on boundary values defined in parameters $\Theta_{init}$, spanning diverse initial drone-target configurations, with $\bm{d}_{0}$ denoting the normalized target direction vector. A trajectory library with occupancy relations, conditioned on the current state, is regenerated at the beginning of each replanning step (Lines 8-12, detailed in Section \ref{sec:trajectory_library}), eliminating the need for a large-scale precomputed library. {Unsafe primitives are rapidly identified and excluded based on perceived point clouds and preassigned occupancy relations, then the safest one is chosen by a composite cost function (Lines 13-15, detailed in Section \ref{sec:labeling_selection})}. If a valid primitive is found, the corresponding data, containing velocity $\bm{v}_{0}$, acceleration $\bm{a}_{0}$, direction $\bm{d}_{0}$, perceived point clouds $\mathcal{P}$, and polynomial coefficients $\bm{c}^*$ of the selected primitive, are stored in a temporary buffer $\mathcal{D}_{temp}$. The primitive is then partially executed to update the drone state. Only when the drone reaches the target within a predefined tolerance is $\mathcal{D}_{temp}$ appended to the final dataset $\mathcal{D}$. This helps encode drone-target proximity and enables the network to learn precise reaching behavior from the dataset. The process repeats with new maps until the valid data number $n_{valid}$ reaches $n_{max}$, after which the full dataset $\mathcal{D}$ is returned.

\subsubsection{Trajectory Library with Occupancy Relations}
\label{sec:trajectory_library}
The library $\mathcal{L}$ is constructed via state lattice discretization, with state dimensions extending to acceleration. Since the initial state $\bm{s}_{0}$ is already known at the replanning step, only terminal states $\bm{s}_{f}$ need to be sampled (Line 9), which significantly reduces the library size. The offline process further enables multi-constraint optimization to generate a set of candidate primitives that are both dynamically feasible and spatially diverse without considering latency.

The trajectory primitive {$\bm{\xi}(t)\in\mathbb{R}^{3}$} is parameterized as a single-segment polynomial vector function in three-dimensional space:
{
\begin{align}
    \bm{\xi}(t) &= [x(t),y(t),z(t)]^T\in\mathbb{R}^{3}=\bm{c}^T\bm{\beta}(t), \label{eq:primitive}\\
    \text{where}\quad\bm{c} &= [\bm{c}_x \ \bm{c}_y \ \bm{c}_z]\in\mathbb{R}^{(n+1)\times3}, \nonumber\\
    \bm{c}_{\mu} &= [c_{n\mu}, \ldots, c_{2\mu}, c_{1\mu}, c_{0\mu}]^T\in\mathbb{R}^{n+1},\;\mu \in \{x, y, z\}, \nonumber\\
    \bm{\beta}(t) &= [t^n,\dots,t^2,t,1]^T\in\mathbb{R}^{n+1},\ t \in [0,\tau]. \nonumber
\end{align}}%
here $x(t)$, $y(t)$ and $z(t)$ are the drone positions; $n$ is the polynomial order; $\bm{c}\in\mathbb{R}^{(n+1)\times3}$ is the coefficient matrix; and {$\tau$ is the fixed time duration}. 

We solve for the primitive coefficients with minimum control effort. {Let $\bar{\bm{c}} = \text{vec}(\bm{c}) \in \mathbb{R}^{3(n+1)}$ denote the column-wise vectorization of $\bm{c}$.} Owing to the differential flatness of quadrotors \cite{mellinger2011minimum}, the optimization problem can be formulated as the convex quadratic program (QP) below:
{
\begin{alignat}{2}
    & \underset{\bar{\bm{c}}}{\text{min}}  &\quad& \bar{\bm{c}}^T \bm{Q} \bar{\bm{c}}, \label{eq:QP_a} \\
    & \vphantom{\underset{\bar{\bm{c}}}{\text{min}}}\text{s.t.} &\quad& \bm{A} \bar{\bm{c}} = \bm{b}, \label{eq:QP_b} \\
    &                 &\quad& \bm{G} \bar{\bm{c}} \leq \bm{h}. \label{eq:QP_c}
\end{alignat}
where $\bm{Q}\in\mathbb{R}^{3(n+1)\times3(n+1)}$ is the positive semidefinite quadratic cost matrix; \eqref{eq:QP_b} \& \eqref{eq:QP_c} represent the boundary and dynamical constraints, respectively, with limits specified in parameters $\Theta_{lib}$. Specifically, $\bm{A}\in\mathbb{R}^{6 N_b\times3(n+1)}$ and $\bm{b}\in\mathbb{R}^{6N_b}$ are the boundary constraint matrix and vector, with $N_b$ denoting the number of derivative orders enforced at boundary points, satisfying $n=2N_b-1$; $\bm{G}\in\mathbb{R}^{6(N_b-1)N_s\times3(n+1)}$ and $\bm{h}\in\mathbb{R}^{6(N_b-1)N_s}$ are the dynamical constraint matrix and vector, with $N_s$ denoting the number of uniformly sampled time points for dynamical limit checking. This QP (Line 10) is solved employing OSQP solver \cite{stellato2020osqp}.} {In this work, we consider minimum-jerk trajectories with $n=5$, $N_b=3$ (constraining position, velocity, and acceleration), $N_s=20$,} {and $\tau=2$s.} {The boundary constraints on start states are taken from the current drone state at each replanning step, while the terminal ones are sampled from the end-state lattice.}

The primitives and all other stored data are expressed in a velocity-aligned frame $\mathcal{F}_V$, where the $x$-axis is aligned with the drone's velocity vector, the $y$-axis points along the cross product of the $x$-axis and gravity vector $\bm{g}$, and the $z$-axis completes the right-handed coordinate system. All primitives share the same origin at zero position in this frame. Inspired by \cite{hou2025primitive}, we additionally construct occupancy relations in this stage to enable subsequent fast collision checking. In $\mathcal{F}_V$, a bounding box enclosing all primitives is discretized into small voxels and organized as a \textit{kd}-tree. For each primitive, the indices of all traversed voxels within an inflated collision radius are retrieved and stored in a \textit{hash}-based set, which defines its occupancy relations $\mathcal{O}$ (Line 11), representing the spatial region occupied by the inflated primitive. Binding these relations directly to the corresponding primitive also helps form an implicit primitive-obstacle relationship, benefiting subsequent network learning. The complete trajectory library $\mathcal{L}$ comprises all optimized primitives along with their associated occupancy relations (Line 12).

\subsubsection{Collision Labeling and Best Primitive Selection}
\label{sec:labeling_selection}
The perceived point clouds are downsampled to a fixed size and mapped into the bounding box to obtain the corresponding voxel indices (Line 13), where $R^V_W$ denotes the rotation matrix from the world frame $\mathcal{F}_W$ to frame $\mathcal{F}_V$. A primitive is labeled unsafe and removed from the library if any of these indices are contained in its occupancy relations (Line 14). This process is highly efficient, as voxel inclusion is checked via \textit{hash} lookup, and a primitive is marked unsafe once the first matching voxel is found. Moreover, it eliminates frequent Euclidean Signed Distance Field (ESDF) queries required by existing learning-based methods during data collection or training \cite{han2025hierarchically,lu2023lpnet,lu2024you,han2025dyna}, making it resource-friendly as well.

The best {primitive} is selected from the remaining safe {candidates} via a task-oriented cost function (Line 15):
\begin{equation}
    \mathcal{C} = w_{t}\mathcal{C}_{target}+w_{l}\mathcal{C}_{length}.
\end{equation}
where $w_t$ and $w_l$ are the weights; $\mathcal{C}_{target}$ and $\mathcal{C}_{length}$ are the target-approach cost and primitive-length cost, respectively, calculated by:
\begin{equation}
\resizebox{0.9\linewidth}{!}{$
    \mathcal{C}_{target} = 
    \begin{cases}
        \|\bm{\xi}(\tau) - \bm{p}_{target}^{\mathcal{F}_V}\|, \quad \text{if } \|\bm{p}_{target}^{\mathcal{F}_V}\| > \|\bm{\xi}(\tau)\|, \\
        \underset{i=0,\dots,N}{\text{min}} \|\bm{\xi}(t_i) - \bm{p}_{target}^{\mathcal{F}_V}\|, \quad \text{otherwise}.
    \end{cases}
$}
\end{equation}
\begin{equation}
    \mathcal{C}_{length} = \sum_{i=0}^{N-1} \|\bm{\xi}(t_{i+1}) - \bm{\xi}(t_i)\|.
\end{equation}
where $\bm{p}_{target}^{\mathcal{F}_V}$ is the target position vector in frame $\mathcal{F}_V$; $t_i=\frac{i\tau}{N}$, with $N$ being the discrete sample number for the primitive.

{Note that $\mathcal{C}_{target}$ encodes the primary objective of target reaching. Its piecewise formulation promotes forward progress when the target lies beyond the primitive's reachable range, while preventing overshoot to facilitate precise target acquisition when the target is within reach. Meanwhile, $\mathcal{C}_{length}$ serves as a geometric regularizer that penalizes unnecessarily curved or weaving primitives and favors more direct trajectories.}

{We clarify the optimality of the expert strategy at two levels. a) \textit{Individual-primitive level}: each candidate primitive is the optimal solution of a constrained minimum-jerk QP under its conditioned boundary states, dynamical limits, and fixed duration. It is optimal in the minimum-control-effort sense, not in the time-optimal sense. b) \textit{Library-selection level}: the expert strategy pursues spatial optimality through a step-wise pipeline of free-space sampling, collision removal, and optimal selection. This design decouples obstacle avoidance from goal reaching, avoiding the trade-offs of jointly optimizing both in a single objective, and thus yields spatially near-straight expert trajectories.}

\subsection{Fast End-to-end Trajectory Planning Framework}

We expect to perform fast online inference of executable trajectories relying solely on onboard sensors. To this end, we consider the following characteristics: a) \textit{End-to-end policy}: Unifying the mapping, front-end path searching, and back-end trajectory solving within the classical planning framework into an integrated process, thereby eliminating inter-module latency and compounding errors; b) \textit{Executable outputs}: {Incorporating state continuity, target-reaching progress quality, and empirical obstacle avoidance and dynamical feasibility for direct execution by low-level controller;} c) \textit{Lightweight network}: With low inference computation and memory consumption, enabling efficient real-time onboard deployment. The framework consists of the following four components:

\subsubsection{Policy Input-Output Design}
To eliminate the need for back-end trajectory solving, the policy is intended to directly output the polynomial coefficients. For a minimum-jerk primitive expressed in frame $\mathcal{F}_V$, the coefficient matrix in \eqref{eq:primitive} can be expanded as:
\begin{equation}
\resizebox{0.9\linewidth}{!}{$
    \bm{c}=
    \begin{bmatrix}
        \bm{c}_a \\[2pt]
        \bm{c}_b
    \end{bmatrix},\;
    \bm{c}_a=
    \begin{bmatrix}
        c_{5x} & c_{5y} & c_{5z} \\
        c_{4x} & c_{4y} & c_{4z} \\
        c_{3x} & c_{3y} & c_{3z}
    \end{bmatrix},\;
    \bm{c}_b=
    \begin{bmatrix}
        \tfrac{1}{2}a_{0x} & \tfrac{1}{2}a_{0y} & \tfrac{1}{2}a_{0z} \\
        \|\bm{v}_0\| & 0 & 0 \\
        0 & 0 & 0
    \end{bmatrix}.
$}
\end{equation}
It can be seen that $\bm{c}_b\in\mathbb{R}^{3\times3}$ governs state continuity and can be directly determined from the current velocity $\bm{v}_0$ and acceleration $\bm{a}_0$. In contrast, $\bm{c}_a\in\mathbb{R}^{3\times3}$ needs to be learned to satisfy obstacle avoidance and target-reaching progress quality, which are associated with the perceived point clouds $\mathcal{P}$ and the target direction $\bm{d}_0$, respectively. By imitating expert coefficients that are carefully selected into the training dataset, {the empirical dynamical feasibility of the inferred trajectories can also be reinforced}. Therefore, the policy network is designed to take $\|\bm{v}_0\|\in\mathbb{R}^1$, $\bm{a}_0\in\mathbb{R}^3$, $\bm{d}_0\in\mathbb{R}^3$ and $\mathcal{P}\in\mathbb{R}^{m\times3}$ as inputs and output $\bm{c}_a\in\mathbb{R}^9$, where $\bm{c}_a$ is expanded into a vector here and $m$ denotes the number of 3D point clouds, thereby enabling end-to-end generation of executable trajectories from sensory data without any further solving.

\subsubsection{Network Architecture}
We present \textit{Npe2eNet}, shown in Fig. \ref{fig:System overview}B, which primarily comprises a point cloud encoder, a feature fusion module, and a coefficient prediction head. {The encoder plays a pivotal role in capturing environmental context. It first pre-processes arbitrary-sized point clouds to a fixed $m_{fix}$ points (set to 666 to balance input size and perceptual sufficiency). Each point's 3D coordinates pass through an MLP ([128, 128, 64], ReLU), followed by max-pooling to a 64-D global descriptor, which is refined by another MLP ([64, 32], ReLU) to a 32-D representation. This is concatenated with the 7-D state into a 39-D input to a feature fusion MLP ([256, 512, 256], ReLU), yielding a compact yet expressive feature vector that encapsulates environmental, dynamical and target information. This feature finally feeds a coefficient prediction MLP ([128, 128, 9], ReLU) to output the nine polynomial coefficients.}

\textit{Npe2eNet} adopts a streamlined architecture that avoids complex operations like convolutions. It can be calculated that the network only contains approximately 0.355M parameters with a computational cost of approximately 34M FLOPs. {The model occupies about 1.38MiB of storage. During online inference, the per-forward-pass consumes about 0.65MiB peak incremental memory and 11.49MiB  peak total memory of GPU\footnote{{Peak incremental memory measures extra dynamic GPU memory for inference operation only, while peak total memory measures the overall dynamic GPU memory needed for inference, including model parameters, input tensors, intermediates, etc.}}.} Compared with existing learning-based methods built on heavier architectures such as ResNet\cite{lu2023lpnet,lu2024you}, MobileNet\cite{loquercio2021learning}, and Transformer\cite{han2025hierarchically}, its resource demands are substantially lower, enabling fast and efficient real-time inference.

\subsubsection{Training Method}
In real flights, perceived point clouds vary significantly in density and spatial distribution across scenarios, and drone state measurements are inherently noisy. These discrepancies, arising from sensory mismatch, primarily constitute the sim-to-real gap in the end-to-end planning scheme. To bridge this gap and enhance the generalization capability of the simulation-trained policy for zero-shot transfer, we implement two strategies during training: 1) \textit{Point cloud pre-processing}: Considering that real flight point clouds may deviate from the fixed network input size $m_{fix}$ and the simulation dataset cannot cover all variations, {we resample the dataset point clouds before formal training.} For each data sample, with probability $p_{pro}$, the point cloud is downsampled by a factor $r$ of its original size, where $r$ is randomly selected from $[r_{min},1]$ and $0 \leq r_{min} \leq 1$. If the resulting size is smaller than $m_{fix}$, it is padded with points whose coordinates are assigned large values (set as 100), denoting distant obstacles; otherwise, it is randomly downsampled to $m_{fix}$. This broadens density coverage and mitigates unseen scenes during training. Note that in online inference stage, the input point clouds are directly downsampled or padded to $m_{fix}$. 2) \textit{Domain randomization}: Gaussian noise is injected into both the input state values and the coordinates of the pre-processed point cloud, with standard deviations set to proportions $r_{\sigma_s}$ and $r_{\sigma_p}$ of their respective magnitudes. This further improves the robustness of the policy network against sensory noise.

\subsubsection{Loss Function}

The network is trained to minimize the Mean Squared Error (MSE) loss between {expert coefficients $\bm{c}_a^*$ and predicted values $\hat{\bm{c}}_a$:
\begin{equation}
    \mathcal{L}_{\mathrm{coeff}} = \mathrm{MSE}(\bm{c}_a^*,\,\hat{\bm{c}}_a) 
    = \frac{1}{M}\sum_{k=1}^{M}(c_{a,k}^* - \hat{c}_{a,k})^2.
\end{equation}
where subscript $k$ indexes the $M$ elements of 
$\bm{c}_a\in\mathbb{R}^{3\times 3}$, with $M=9$.} AdamW optimizer \cite{loshchilov2017decoupled} (learning rate $lr=1.0\times10^{-3}$, weight decay $=1.0\times10^{-4}$) with cosine annealing scheduling (half-cycle length $T_{max}=0.3\times$maximum epochs, minimum learning rate $\eta_{min}=0.1\times lr$) is utilized to ensure fast, stable convergence and mitigate overfitting. Training employs a batch size of 128 (validation 64), with a maximum of 600 epochs and early stopping patience of 20.

\section{Evaluations}
\label{sec:evaluations}

\subsection{Implementation Details}
We conduct dataset collection, network training, and simulation tests on a desktop computer with i7-10700K CPU, GTX 1060 GPU and Ubuntu 20.04 system. The randomized simulation maps for dataset collection consist mainly of cylinders, rectangular columns, and rings, with overall obstacle densities ranging from 1/40 to 1/4. A total of one million valid samples are collected, with each requiring only approximately 0.08s to generate, and the data are stored in H5 format for training the network via PyTorch. The resulting policy model is deployed to ROS via LibTorch for online inference. In real-world experiments, we implement the learned policy model on a quadrotor without any further fine-tuning. The platform integrates a Livox Mid-360 LiDAR\footnote{\url{https://www.livoxtech.com/mid-360}} for onboard sensing, an NVIDIA Jetson Orin NX\footnote{\url{https://www.nvidia.com/en-us/autonomous-machines/embedded-systems/jetson-orin/}} for end-to-end planning and SE(3) geometric tracking control\cite{lee2010geometric}, and a PX4\footnote{\url{https://px4.io/}} flight controller for low-level attitude control.

{The planned trajectory is executed in an event-triggered receding-horizon manner. Specifically, replanning is triggered by whichever of the following conditions occurs first: the quadrotor travels the preset distance threshold $d_{\mathrm{rep}}=0.5$m, the execution time reaches the upper limit $T_{\mathrm{rep}}^{\max}=2\tau/3$, or a potential collision is detected. This mechanism allows the actual execution time to adapt naturally to the flight speed while remaining within the valid time window $[0,\tau]$. Additionally, each inferred trajectory undergoes a lightweight post-inference collision check \cite{zhou2020ego,zhou2022swarm,ren2025safety} before being sent to the controller. The first two-thirds of the trajectory are sampled at intervals of 0.05s and queried against an inflated occupancy grid via an $\mathit{O}(1)$ lookup. Note that the checker operates in an independent parallel module outside the end-to-end trajectory generation pipeline and serves as a decoupled engineering safeguard against rare inputs outside the training distribution.}

\subsection{Ablation Study}
\label{subsec:ablation}

\begin{figure*}[t] 
    \centering
    \includegraphics[width=\textwidth]{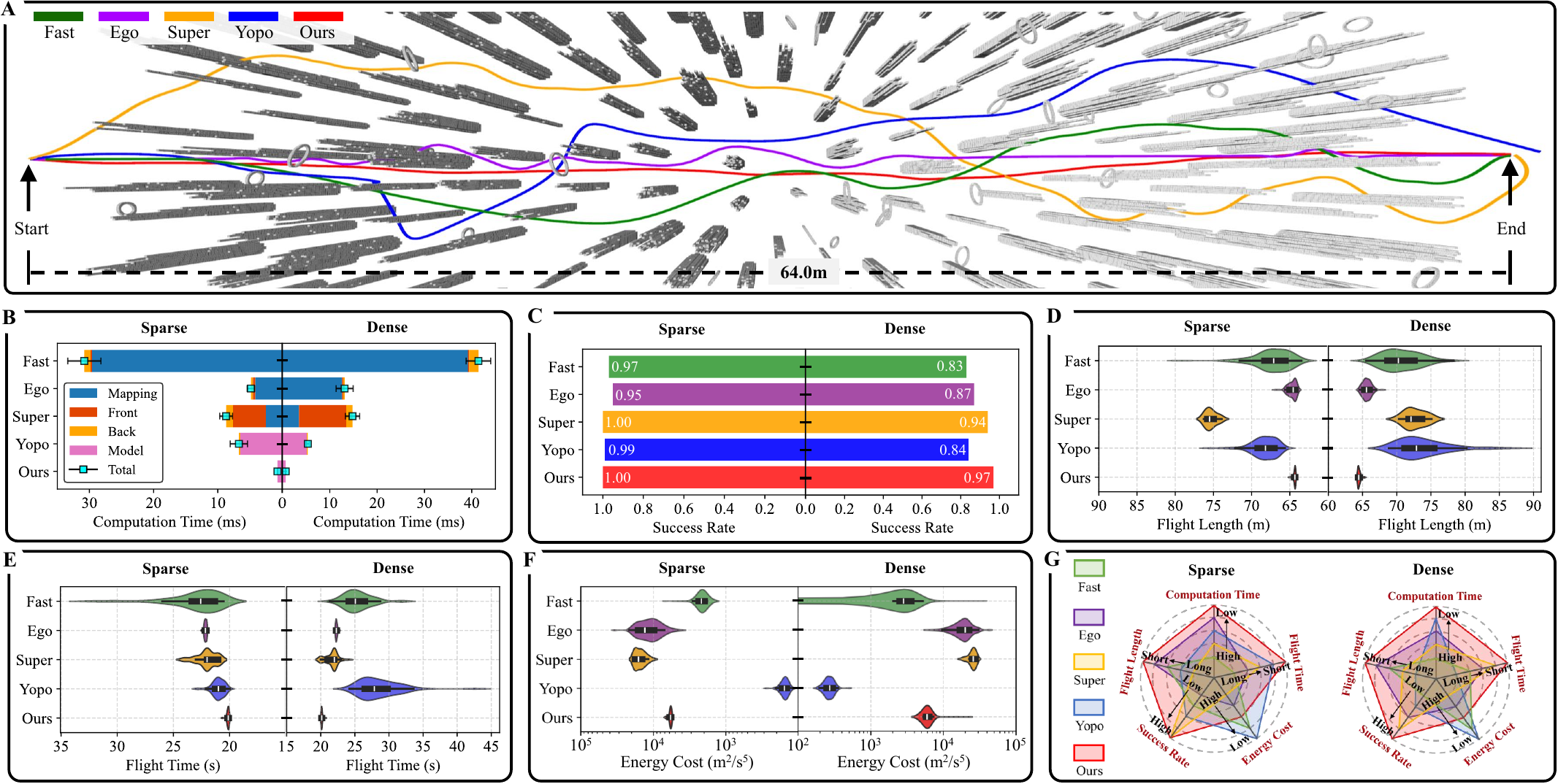}
    \vspace{-0.3cm}
    \caption{\textbf{Benchmark results under different obstacle densities.} \textbf{(A)} Example (successful) trajectories in dense scenarios. \textbf{(B)} Computation time, where {``}Mapping{''} includes grid map, point cloud map, and ESDF construction, {``}Front{''} denotes path finding and corridor generation, {``}Back{''} indicates trajectory optimization or closed-form BVP solving, and {``}Model{''} refers to network inference. \textbf{(C)} Success rate. \textbf{(D)} Flight length. \textbf{(E)} Flight time. {\textbf{(F)} Energy cost. \textbf{(G)} Overall comparison across the five metrics.}}
    \label{fig:Sim compare}
    \vspace{-0.5cm}
\end{figure*}

We first conduct a series of ablation experiments to examine how the training augmentation techniques affect network performances, with a focus on the key parameters in the domain randomization (i.e., $r_{\sigma_s}$ and $r_{\sigma_p}$) and point cloud pre-processing (i.e., $p_{pro}$ and $r_{min}$) modules.

All experiments are performed in 50$\times$20$\times$10m simulated maps with \textit{dense} obstacle configurations (160 cylinders and 40 rings, minimum spacing 2m, and overall density 1/5). The {drone} autonomously {navigates} a 64m start-target distance, with success defined as reaching the target without collision (drone radius 0.15 m) and within a 1m target tolerance. The maximum speed is limited to 4m/s. Each ablation setting is evaluated over 250 runs. In every run, {obstacle} positions and radii are randomly generated{. Gaussian noise with zero mean and} a 5\% standard deviation is injected into both the network input states and point cloud coordinates to emulate measurement uncertainties. {The results are reported in Table \ref{table:ablation_A} and Table \ref{table:ablation_B}.}

\begin{table}[t]
  \renewcommand{\arraystretch}{1.3}
  \centering
  \begin{threeparttable}
  \caption{{Ablation study on domain randomization parameters}}
  \label{table:ablation_A}
  \begin{tabular*}{\columnwidth}{@{\extracolsep{\fill}}cccc}
  \toprule
  No. & $r_{\sigma_s}$ & $r_{\sigma_p}$ & Success Rate \\
  \midrule
  A0 & 0.00 & 0.00 & 0.904 \\
  A1 & 0.03 & 0.03 & 0.924 \\
  A2 & 0.03 & 0.04 & \textbf{0.936} \\
  A3 & 0.04 & 0.04 & 0.912 \\
  A4 & 0.06 & 0.06 & 0.892 \\
  \bottomrule
  \end{tabular*}
  \begin{tablenotes}
    \footnotesize
    \item {Note: conducted with the point cloud pre-processing module disabled}
  \end{tablenotes}
  \end{threeparttable}
\end{table}

\begin{table}[t]
  \renewcommand{\arraystretch}{1.3}
  \centering
  \begin{threeparttable}
  \caption{{Ablation study on point cloud pre-processing parameters}}
  \label{table:ablation_B}
  \begin{tabular*}{\columnwidth}{@{\extracolsep{\fill}}cccc}
  \toprule
  No. & $p_{pro}$ & $r_{min}$ & Success Rate \\
  \midrule
  B0 & 0.0 & 1.0 & 0.936 \\
  B1 & 0.3 & 0.7 & 0.944 \\
  B2 & 0.6 & 0.5 & \textbf{0.972} \\
  B3 & 0.6 & 0.3 & 0.932 \\
  B4 & 0.9 & 0.1 & 0.888 \\
  \bottomrule
  \end{tabular*}
  \begin{tablenotes}
    \footnotesize
    \item {Note: conducted on top of the selected A2 domain randomization setting}
  \end{tablenotes}
  \end{threeparttable}
\end{table}

\begin{table}[t]
	\renewcommand{\arraystretch}{1.3}
	\centering
	\begin{threeparttable}
		\caption{{Sensitivity analysis on point cloud input size}}
		\label{table:ablation_m_fix}
		\setlength{\tabcolsep}{1.5pt}
		\begin{tabular*}{\columnwidth}{@{\extracolsep{\fill}}ccccc}
			\toprule
			$m_{fix}$ & \makecell{Success\\Rate} & \makecell{Inference\\Latency (ms)} & \makecell{Peak Incremental\\Memory (MiB)} & \makecell{Peak Total\\Memory (MiB)} \\
			\midrule
			333  & 0.916 & 0.43 & 0.325 & 11.162 \\
			666  & 0.972 & 0.67 & 0.650 & 11.491 \\
			1024 & 0.968 & 0.78 & 1.003 & 11.844 \\
			2048 & 0.976 & 0.93 & 2.001 & 12.856 \\
			\bottomrule
		\end{tabular*}
	\end{threeparttable}
\end{table}

Experiments A0 to A4 assess different ($r_{\sigma_s}$, $r_{\sigma_p}$) combinations under domain randomization, with the point cloud pre-processing module disabled. Compared with the baseline A0, success rates are observed to improve in A1 to A3 when noise is added to state values and point cloud coordinates during training. These perturbations enhance robustness to proprioceptive and exteroceptive disturbances, respectively, and {mitigate} overfitting to idealized clean observations. However, excessive noise, as in A4, degrades performance due to distortion of the original observation distribution and impairment of stable input-output learning, leading to potential misinterpretation of obstacles and targets. By evaluating both symmetric and asymmetric combinations, we select A2 ($r_{\sigma_s}$=0.03, $r_{\sigma_p}$=0.04, and success rate 0.936) as the domain randomization setting.

Experiments B0 to B4 further investigate the impact of different ($p_{pro}$, $r_{min}$) combinations with the point cloud pre-processing module added on top of domain randomization, where A2 serves as the baseline B0. For autonomous flight in clutters, point clouds encode critical obstacle constraints for safe navigation. Previous noise injection on point cloud coordinates essentially perturbs spatial positions only, but overlooks variations in point cloud density. The proposed pre-processing module exposes the network to a broader density range during training, enhancing adaptation to fluctuations in obstacle point counts. Consequently, success rates are further improved (as in B1 and B2) over the domain randomization baseline B0 in complex scenarios. Moreover, $p_{pro}$ governs the pre-processing frequency, while $r_{min}$ sets the lower bound for point cloud density. Insufficient $p_{pro}$ or excessive $r_{min}$ fails to capture density fluctuations adequately, leading to marginal robustness improvements in B1. Conversely, excessively high $p_{pro}$ paired with minimal $r_{min}$ causes over-sparsification, eroding the obstacle geometric fidelity within the dataset and in turn compromising success rates in B3 and B4. The balanced configuration B2, with $p_{pro}$ = 0.6 and $r_{min}$ = 0.5, achieves a peak success rate of 0.972. These results verify the efficacy of the proposed point cloud pre-processing strategy.

{We further conduct a sensitivity analysis on the fixed point cloud input size $m_{fix}$, examining its effects on success rate, inference latency, peak incremental memory, and peak total memory. The results are reported in Table \ref{table:ablation_m_fix}. It can be seen that reducing $m_{fix}$ to 333 lowers latency and memory consumption, which is reasonable since the point cloud encoder operates in a point-wise manner, making latency and dynamic memory scale with input size. However, the success rate in case 333 drops to 0.916, indicating insufficient environmental information for reliable planning compared to case 666. Moreover, increasing $m_{fix}$ to 1024 and 2048 yields only marginal changes in success rate, but incurs higher latency and memory costs, with peak incremental memory rising by 54.3\% and 207.8\%, respectively. These results indicate that case 666 offers a favorable balance among planning reliability, computational efficiency, and memory cost.}

\subsection{Acceleration Continuity Analysis}

\begin{figure}[t]
    \begin{center}
        \includegraphics[width=\columnwidth]{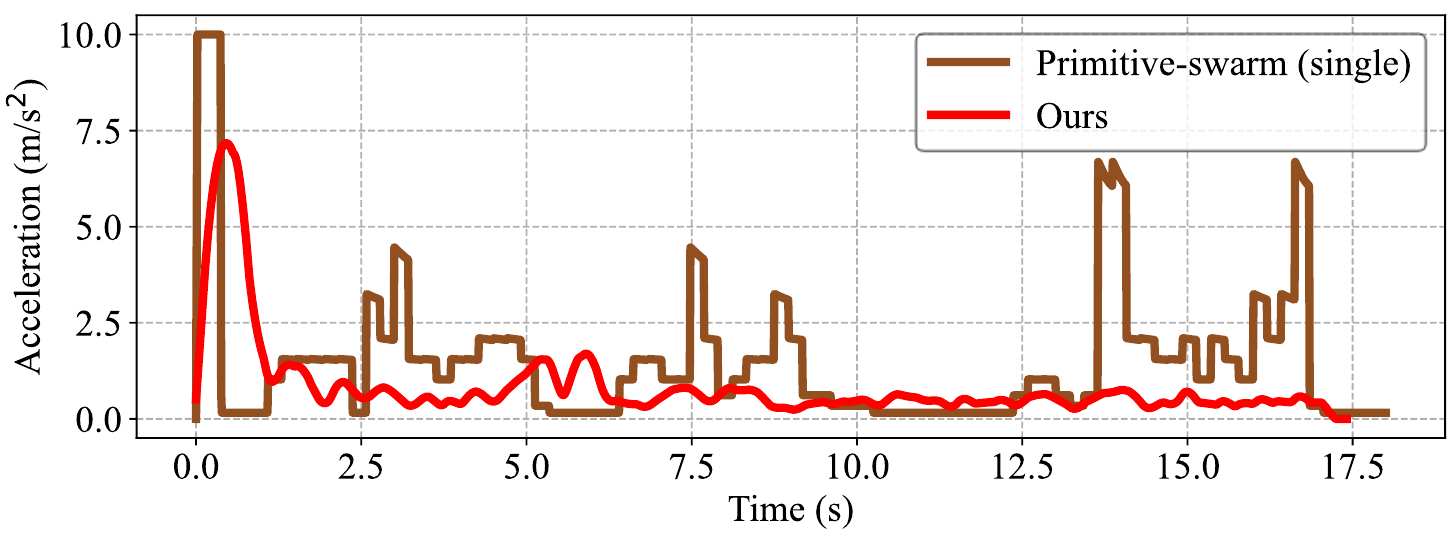}
    \end{center}
    \vspace{-0.3cm}
    \caption{\textbf{Acceleration curves.} Our method overcomes the high-order state discontinuity problem of traditional offline library-based planners \cite{hou2025primitive}.}
    \label{fig:acc_discontinuity}
    \vspace{-0.5cm}
\end{figure}

To demonstrate the resolution of the acceleration discontinuity inherent in traditional offline primitive library-based planners, we compare with the single version of \textit{Primitive-swarm} \cite{hou2025primitive} and plot the curves (Fig. \ref{fig:acc_discontinuity}) under \textit{dense} environments. It is evident that acceleration jumps occur throughout the entire flight of \cite{hou2025primitive}. {This occurs because the library cannot cover all possible acceleration states due to size constraints for onboard memory and latency}. As a result, when an uncovered initial acceleration arises during online selection, discontinuities appear, degrading trajectory quality and increasing motor load. In contrast, our method explicitly considers higher-order state continuity when designing the network input and output, thereby successfully addressing this problem.

\subsection{Benchmarks}

In this section, we compare the proposed method with four representative local planners, including three well-established approaches under the classical hierarchical planning framework and one state-of-the-art learning-based planner, namely: 1) \textit{Fast}\cite{zhou2019robust}: Kinodynamic path searching on an incrementally updated voxel map, followed by gradient-based trajectory optimization relying on a constantly maintained ESDF for safety regularization; 2) \textit{Ego}\cite{zhou2020ego}: A$^*$ path searching on the constructed grid map, with subsequent ESDF-free trajectory optimization leveraging collision costs derived directly from grid-obstacle information; 3) \textit{Super}\cite{ren2025safety}: Path searching and flight corridor generation directly on an efficient spatiotemporal sliding point cloud map, followed by MINCO-based \cite{wang2022geometrically} dual-trajectory optimization that jointly accounts for high-speed exploration and safety assurance; 4) \textit{Yopo}\cite{lu2024you}: Objective-function-guided unsupervised learning for the network, inferring the end-state related terms of all pre-sampled primitives from sensory inputs, after which one is selected for closed-form BVP solving. 

All planners are tested in simulated maps with different obstacle densities. The autonomous navigation task and \textit{dense} environment configuration follow those described in Sect. \ref{subsec:ablation}. The \textit{sparse} configuration consists of 40 cylinders and 10 rings, with a minimum spacing {of} 4m and an overall density of 1/20. {All baselines are evaluated with their official open-source implementations and default parameters. For a fair comparison, all planners use the same 90$^\circ$$\times$60$^\circ$ Field-of-View (FOV), collision radius of 0.15m, maximum velocity of 4m/s, and simulator (including the controller and quadrotor model) adopted from \cite{zhou2019robust}.} The target reaching tolerance is still set {to} 1m, but relaxed to 4m for \textit{Yopo}, as it cannot satisfy stricter tolerances. Each planner is evaluated over 100 runs in both dense and sparse scenarios, with obstacles randomly generated in each run. Flight visualizations and statistical results are presented in Fig. \ref{fig:Sim compare}.

It can be seen that our method outperforms the baselines in both sparse and dense environments (Fig. \ref{fig:Sim compare}{G}), whereas the baselines, despite performing well in sparse settings, deteriorate in dense scenarios. In terms of computation time (Fig. \ref{fig:Sim compare}B), \textit{Fast} incurs the highest cost, primarily due to ESDF maintenance. {Although \textit{Ego} and \textit{Super} avoid this overhead, they still require grid map construction or point cloud-based corridor generation, so their planning time is highly sensitive to obstacle density and increases with it.} \textit{Yopo} circumvents mapping and front-end processing, running faster than the first three methods in dense environments. However, in sparse settings, its runtime even slightly exceeds \textit{Ego} due to the need {for} inference over all pre-sampled primitives, complex in-network operations, and back-end {BVP solving} costs. Consequently, when grid map construction exerts less influence in sparse {environments}, \textit{Yopo} loses its efficiency advantage over \textit{Ego}. Our method achieves the shortest computation time, averaging 0.70ms (sparse) and 0.68ms (dense), about 7 to 60 times faster than the baselines, with stable efficiency independent of densities due to its highly integrated end-to-end planning policy and compact yet effective architecture. 

Regarding task-oriented trajectory optimality (Fig. \ref{fig:Sim compare}D and Fig. \ref{fig:Sim compare}E), the four baselines yield longer and slower trajectories than the proposed method. This arises from their strong coupling of obstacle avoidance and goal navigation into a complex multi-constraint optimization problem (including \textit{Yopo}, which essentially formulates the same problem but solves it with a neural network). As a result, they must trade off among competing factors, often becoming trapped in local minima and producing detours, as shown in Fig. \ref{fig:Sim compare}A. By contrast, our method benefits from imitating the customized primitive strategy that decouples the two problems by first eliminating colliding primitives and then concentrating on selecting the task-oriented optimal primitive from the safe set, thereby avoiding compromises for obstacle avoidance. This design enables near-direct flight distances (64.51m in dense and 64.39m in sparse) and the shortest flight time to the target. 

{Fig. \ref{fig:Sim compare}F shows the energy cost measured by the integral of squared-jerk. Consistent with the trajectory results, our method yields lower accumulated squared-jerk than \textit{Ego} and \textit{Super}, while the lower values of \textit{Fast} and \textit{Yopo} are achieved at the expense of substantially longer detours and less direct motion toward the target. This indicates that our method achieves a favorable balance between efficient target reaching and energy consumption.} Finally, while the other planners exhibit declining success rates in dense environments, our method sustains the highest rate of 0.97 (Fig. \ref{fig:Sim compare}C). This stems from its ultra-low latency, enabling rapid responses to dense obstacles even at high speeds, further reinforced by point-cloud pre-processing and domain randomization for robust environmental abstraction.

\begin{figure}[t]
    \begin{center}
        \includegraphics[width=\columnwidth]{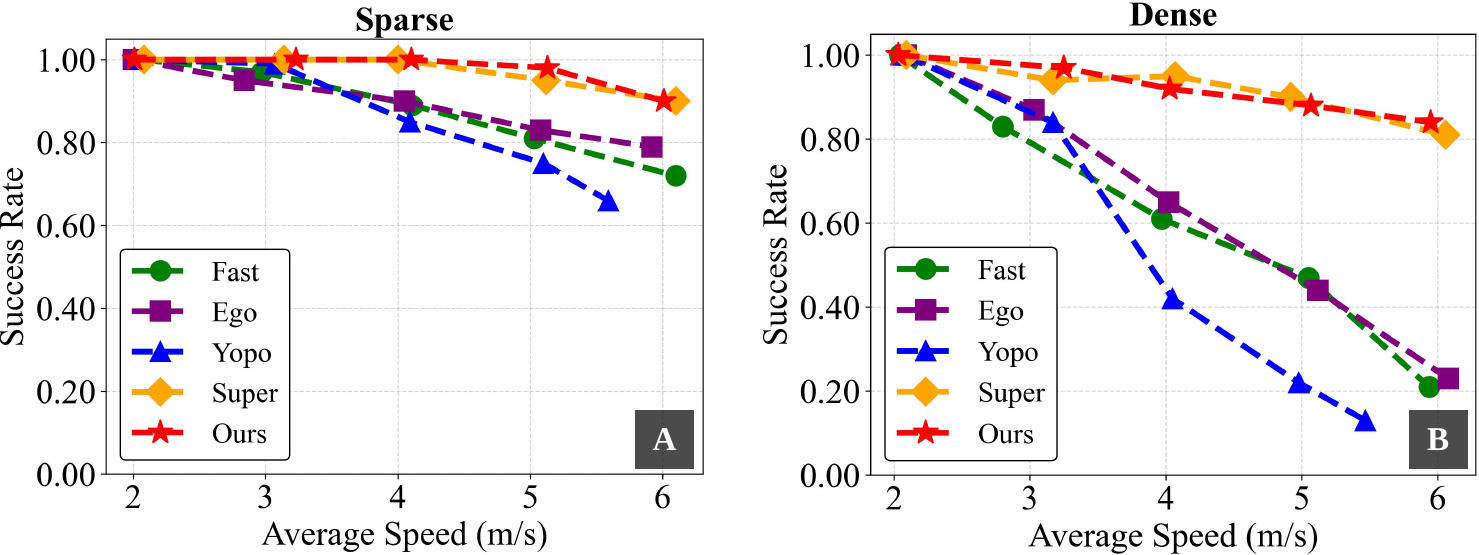}
    \end{center}
    \caption{\textbf{Success rates across average speeds.} \textbf{(A)} Results in sparse environments. \textbf{(B)} Results in dense environments.}
    \label{fig:speeds_success}
    \vspace{-0.5cm}
\end{figure}

Fig. \ref{fig:speeds_success} plots success rates against increasing average speeds for all compared planners. Results indicate that all methods achieve perfect success rates at low speeds ($\approx$2m/s). However, performance degrades significantly for \textit{Fast}, \textit{Ego}, and \textit{Yopo} as speed increases, particularly in dense scenarios. In contrast, \textit{Super} and our proposed method maintain success rates above 0.80 even at high average speeds of around 6m/s. The decline for \textit{Fast} and \textit{Ego} primarily arises from hierarchical latency and compounding errors. Moreover, their joint optimization processes frequently converge to local minima, failing to generate safe trajectories under high speed motion. For \textit{Yopo}, we observe inconsistent velocity profiles characterized by abrupt accelerations, leading to collisions when obstacles appear suddenly. Furthermore, the lack of explicit density-aware training also limits its generalization across varying obstacle configurations. Although \textit{Super} employs a hierarchical planning structure, its complex dual-trajectory mechanism ensures an effective balance between speed and safety. Our method achieves comparable success rates to \textit{Super} but avoids its extensive hand-crafted rules and pronounced trajectory detours as demonstrated in Fig. \ref{fig:Sim compare}A. Through a streamlined end-to-end architecture, our approach produces straighter flight paths while maintaining the same high level of robustness.

\begin{figure*}[t]
    \centering
    \includegraphics[width=\columnwidth]{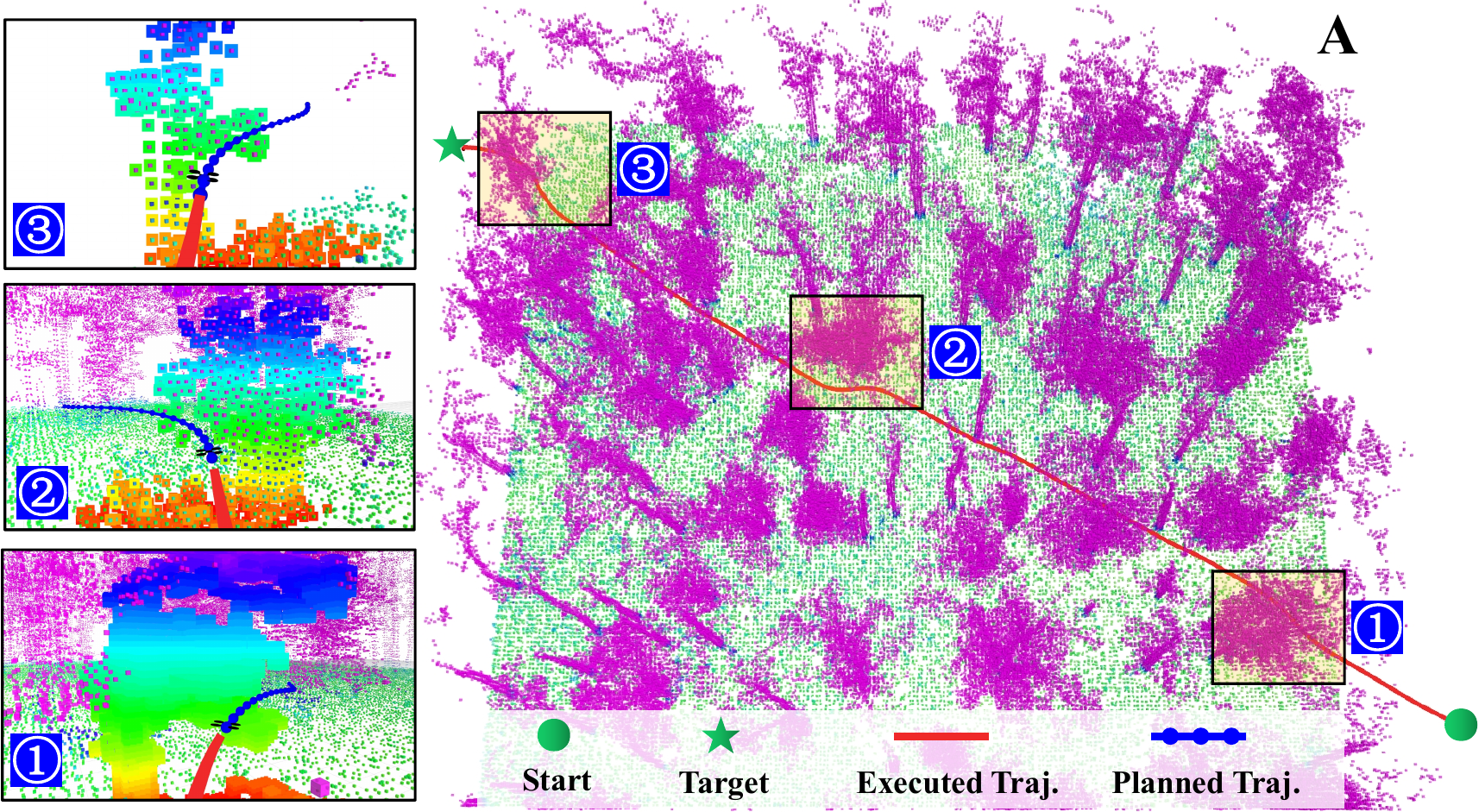}\hfill
    \includegraphics[width=\columnwidth]{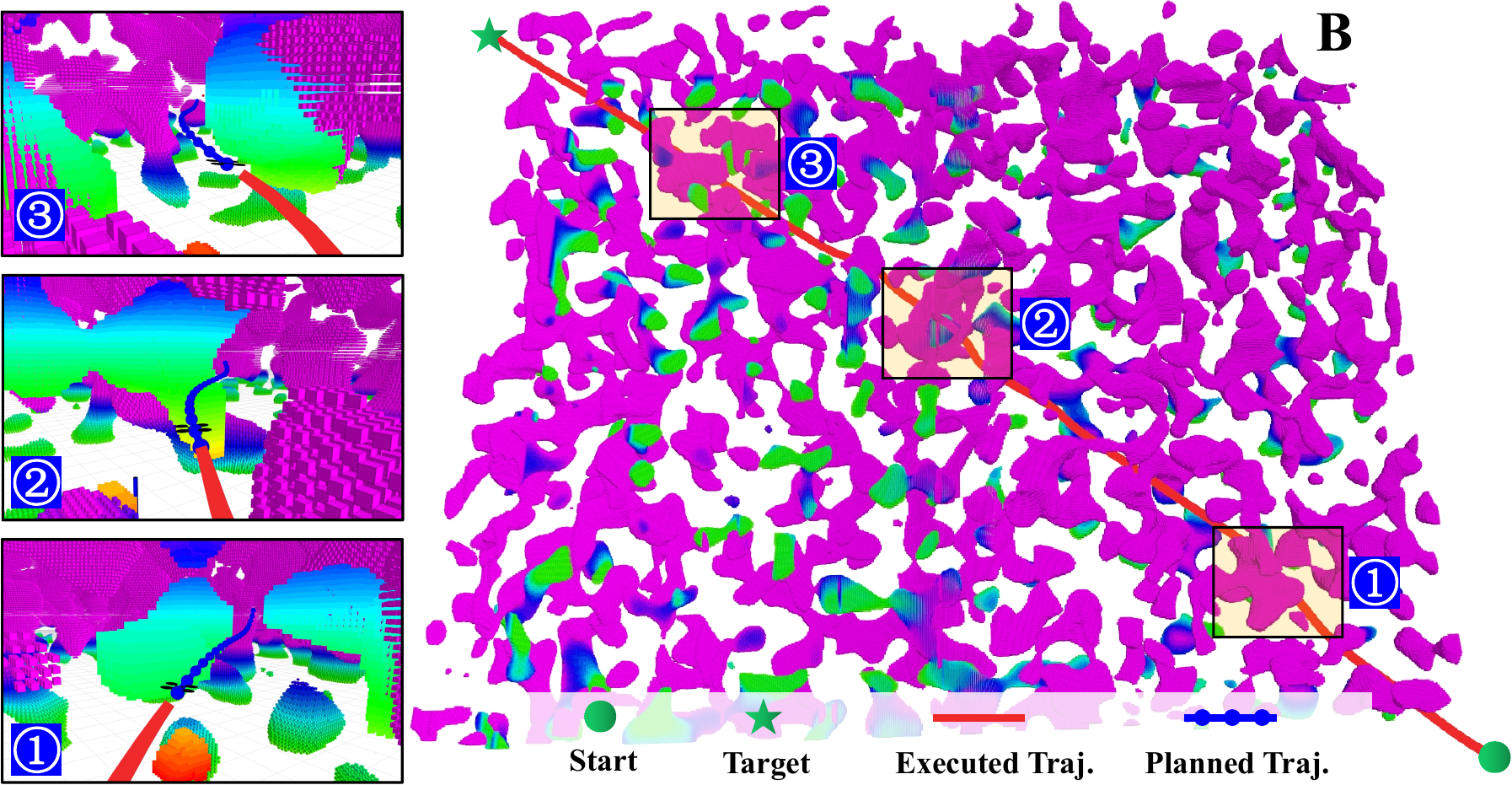}
    \caption{\textbf{Generalization test environments.} The main panels show the global top view, while the adjacent insets present the corresponding first person views of the indexed local regions. \textbf{(A)} Point cloud map collected from real world forests \cite{chaney2023m3ed}. \textbf{(B)} Simulated 3D obstacle map resembling caves and mountainous terrain \cite{han2025dyna}.}
    \label{fig:generalization_combined}
\end{figure*}

\subsection{Generalization Evaluations}

To evaluate the generalization capability of the proposed planner, we test the learned policy via simulation in environments not encountered during training. {Two types of more complex new maps are employed}: 1) an open-sourced point cloud map collected from real-world forests \cite{chaney2023m3ed}, which we crop to a volume of 60$\times$50$\times$20m. This map contains not only tree trunks, branches, and ground vegetation, but also spatially distributed noise points, as illustrated in Fig. \ref{fig:generalization_combined}A; and 2) a simulated unstructured 3D obstacle map resembling caves and mountainous terrain \cite{han2025dyna}, generated using Perlin noise with dimensions of 60$\times$50$\times$10m. In this second map, obstacles exhibit diverse irregular shapes and varying densities that differ from those in the training datasets, as shown in Fig. \ref{fig:generalization_combined}B. For each map type, we perform 150 autonomous traversal trials. Start and target positions are randomly initialized with separation distances of 70$\sim$80m per trial. The policy is the one obtained from ablation experiment B2 in Sect. \ref{subsec:ablation}, without any fine-tuning to the new environments. The resulting success rates are 0.907 and 0.833, respectively. It can be seen that although performance degrades relative to training environments, a common limitation of learning-based methods, the results still remain high and exceed those reported in \cite{han2025dyna}. This robustness is attributed to the synergistic effects of domain randomization and point cloud pre-processing. These strategies improve tolerance to spatial noise while enhancing adaptability to density variations. Consequently, the proposed planner achieves comparable generalization across both noise-corrupted and density-varying new environments.

\subsection{Real-world Experiments}
We validate our method through real-world experiments in both outdoor unknown forests and indoor cluttered environments. All scenarios are previously unseen, and the training data come solely from simulation without real-world fine-tuning. We refer readers to the supplementary video for more information.

As illustrated in Fig. \ref{fig:Real_world_exp}A, we first conduct the outdoor navigation experiments in a non-uniform forest with approximately 1/6 trees/m$^2$ density and 0.3m average tree diameter. Each tree is irregularly surrounded by multiple tilted wooden stakes, forming a complex cluttered environment. The results show that the quadrotor autonomously flies to a target 60m ahead (within 1m reaching tolerance) under random disturbances including wind and sensor noise. It reaches a maximum speed of 6.10m/s, completes the flight within 12s without collisions, and maintains a nearly straight trajectory without significant detours, demonstrating effective sim-to-real transfer capability.

\begin{figure}[t]
    \begin{center}
        \includegraphics[width=\columnwidth]{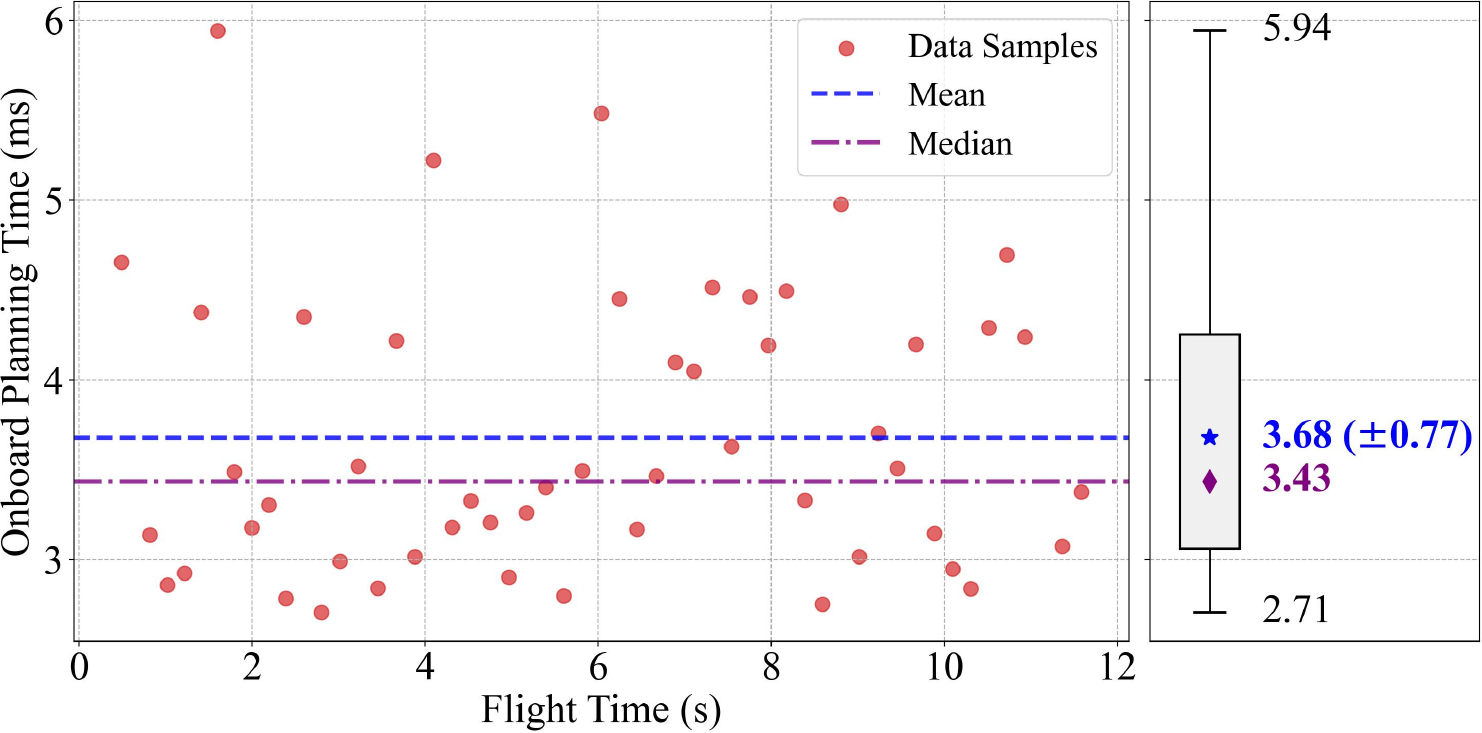}
    \end{center}
    \caption{\textbf{Statistics of onboard planning time.} The distribution characterizes the superior real-time onboard efficiency of the proposed end-to-end local planner with a mean planning time of only 3.68ms throughout the flight.}
    \label{fig:Real_world_Latency}
    \vspace{-0.5cm}
\end{figure}

Fig. \ref{fig:Real_world_Latency} presents the statistics of onboard planning time for end-to-end planning throughout the entire flight. The mean is 3.68ms with a standard deviation of $\pm$0.77ms. The median is as low as 3.43ms, with values ranging from 2.71ms to 5.94ms. These metrics highlight the exceptional computational efficiency of the proposed method for onboard deployment.

\begin{figure}[t]
    \begin{center}
        \includegraphics[width=\columnwidth]{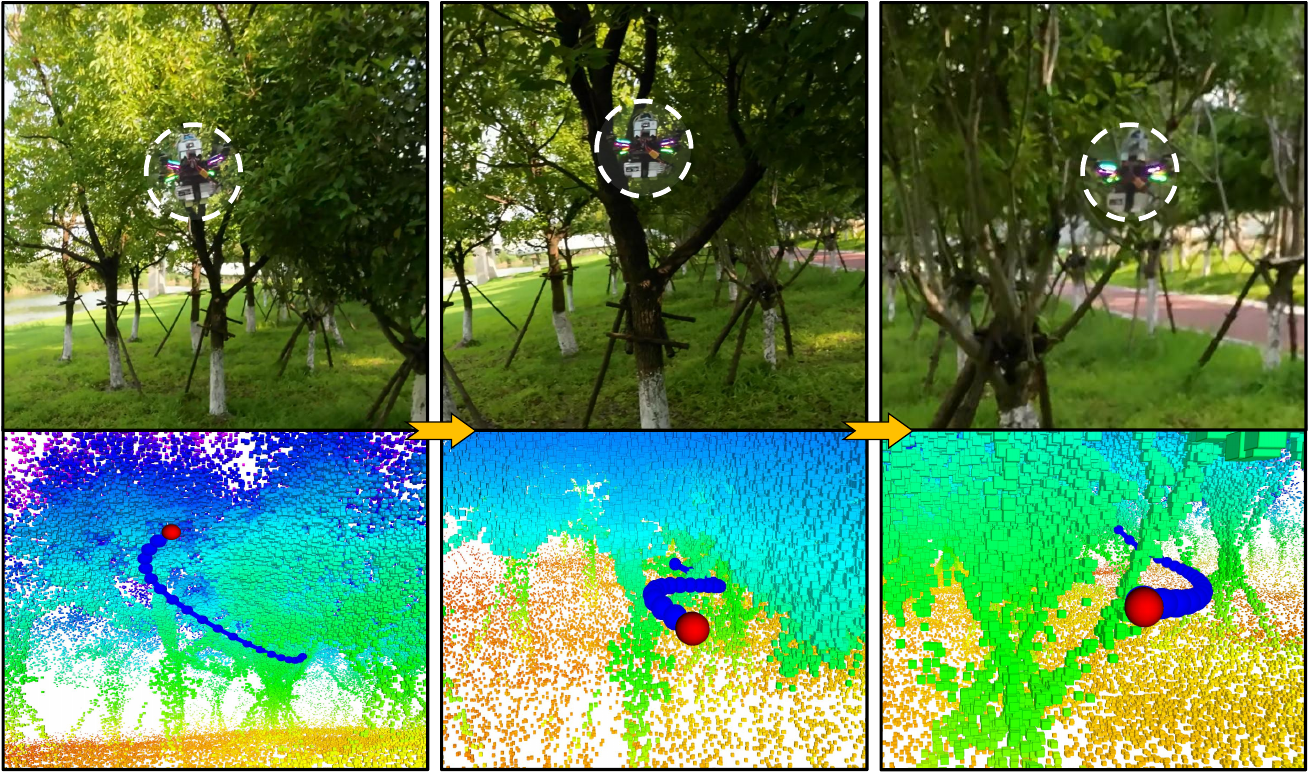}
    \end{center}
    \caption{\textbf{Field snapshots and onboard planning visualizations in thick foliage.} Red spheres and blue curves denote current positions and planned trajectories, respectively. The quadrotor performs safe and robust maneuvering through restricted spaces with a minimum passage under 0.8m.}
    \label{fig:denseForest_exp}
    \vspace{-0.5cm}
\end{figure}

The flight performance in a denser forest with thick foliage is depicted in Fig. \ref{fig:denseForest_exp}. Despite the cluttered obstacles and the narrowest passage under 0.8m (quadrotor diameter $\approx$0.3m), the planner can still successfully generate safe trajectories and execute traversal maneuvers. Furthermore, we also conduct indoor experiments in a confined space with randomly placed obstacles, as shown in Fig. \ref{fig:Real_world_exp}B. The quadrotor needs to sequentially visit multiple preset targets within a 0.5m reaching tolerance before proceeding to the next. This experiment validates the stability and safety of the proposed planner during continuous multi-target navigation in unknown, cluttered environments.

\section{Conclusion and Future Work}
\label{sec:conclusion}

This paper leverages imitation learning to inherit the advantages of offline primitive-based methods in generating high-quality trajectories, while addressing their inherent discontinuity drawback through online network inference. The resulting planner directly outputs polynomial coefficients from sensory inputs{. It produces} {practically} controller-executable trajectories without back-end solving { and forms} a highly integrated end-to-end planning scheme. The proposed point cloud pre-processing technique during training improves success rates across varying obstacle densities and enhances generalization capability. Extensive benchmarks against state-of-the-art baselines in both dense and sparse environments demonstrate consistent superiority in computation time, success rate, flight length, and flight time. Furthermore, zero-shot deployment in indoor and outdoor real-world experiments validates the robustness and efficiency of the proposed planner. {Nevertheless, we acknowledge that the current framework is primarily designed for static obstacles and remains limited in handling highly dynamic obstacles.} {Moreover, as a local planner relying on instantaneous limited Field-of-View inputs, it may struggle to escape large local traps such as dead ends and U-shaped obstacles. Overcoming these limitations will be investigated as our future work.}

\bibliography{npe2eLib}
\end{document}